# Zutu: A Platform for Localization and Navigation of Swarm Robots Using Virtual Grids

Prateek
Department of Networking and Communications
SRM Institute of Science and Technology
Kattankulathur, India
pz6011@srmist.edu.in

Pawan Wadhwani
Department of Computational Intelligence
SRM Institute of Science and Technology
Kattankulathur, India
pw7779@srmist.edu.in

Reshesh Kumar Pathak
Department of Electronics and Communications
SRM Institute of Science and Technology
Kattankulathur, India
rp7107@srmist.edu.in

Mayur Bhosale
Department of Mechanical Engineering
SRM Institute of Science and Technology
Kattankulathur, India
mb7549@srmist.edu.in

Dr. A Helen Victoria
Department of Networking and Communications
SRM Institute of Science and Technology
Kattankulathur, India
helenvia@srmist.edu.in

*Abstract*— **Swarm robots, which are inspired from the way insects behave collectively in order to achieve a common goal, have become a major part of research with applications involving search and rescue, area exploration, surveillance etc. In this paper, we present a swarm of robots that do not require individual extrinsic sensors to sense the environment but instead use a single central camera to locate and map the swarm. The robots can be easily built using readily available components with the main chassis being 3D printed, making the system low-cost, low-maintenance, and easy to replicate. We describe Zutu's hardware and software architecture, the algorithms to map the robots to the real world, and some experiments conducted using four of our robots. Eventually, we conclude the possible applications of our system in research, education, and industries.**

*Keywords*— **AR tags, swarm robotics, Autonomous Robots, Collective behaviour, ROS**

## I. Introduction

In nature, a swarm is a group of insects or animals in which each member individually may have a simple task, but as a whole, it accomplishes a complex task which is generally the common goal [1]. Swarm robotics is the branch of robotics that imitates behaviours and patterns observed in nature such as ants, bees, birds, and fish [2]. Such swarms of robots are gaining popularity in research because of their high adaptability and scalability.

The issue however faced in such swarms is the reliance on the use of extrinsic sensors for the measurement of distance, range, bearing, light, etc. The sensors used in each individual robot require timely calibration and maintenance. The use of sensors in individual robots also leads to a substantial increase in the cost of development when scaling.

Zutu is a platform that involves only a single monocular camera that uses a fiducial marker (AR tags) to determine the 6DoF(Degrees of Freedom) pose of each of the bots as well as the area that they work upon. The accuracy of AR tags depends on the average distance between the camera and the marker. Furthermore, a one-time camera calibration can help increase the accuracy [3].

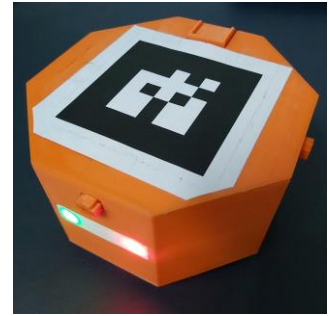

Fig. 1: Prototype of a Zutu robot

The Zutu bot (Fig. 1) measures 15.24 cm x 15.24 cm (6 inches x 6 inches) and is placed on a grid of 304.8 cm x 152.4 cm (120 inches x 60 inches). The chassis is completely 3D printed. The components required are easily available and the monocular camera used is a smartphone camera behaving as an Ip camera to be used wirelessly over the Local Area Network (LAN), further reducing the cost.

The software used in the LAN for interprocess communication is developed using a widely recognized open-source platform for robotics - ROS. ROS helps to easily integrate components with one another and provides easy debugging during development.

**Related work and differentiation.** Swarm robotics has been a huge research field in recent years and a high number of mobile robots have been produced. In this section, we will discuss some of the swarm mobile robots which are relevant to our work, along with their dependencies and capabilities. E-puck [4] is a widely successful mobile robot used for research and education, it uses a large number of sensors such as cameras, microphone arrays, proximity sensors, accelerometers, etc. E-puck uses Bluetooth for communication. AMiR [5] uses an array of IR sensors for environment perception and communication, however. Colias [6] also uses a similar strategy such as AMiR for communication and perception and also overcomes drawbacks of AMiR such as its large size and low speed. Jasmine [7] is another extensively used micro-robot in research that uses six IR sensors to detect obstacles. Jasmine has been used in several BEECLUST scenarios by playing the role of a bee. Zooids [8] is another microrobot measuring only 2.6 cm in diameter. Zooids use a high-speed light projector for optical tracking. Kilobots [9] are small-sized robots that move via vibrating on smooth surfaces, reducing their use on various surfaces. MicroMVP[10] and HeRo [11] use AR tags with a camera for detection and 3D printing technology
HeRo uses proximity sensors for autonomous decisions.

Our system tries to make use of a single central camera for localization and navigation and eliminate or minimise the need to use onboard sensors. Compared to the mentioned counterparts, we use 3D-printed low-cost designs and open-source libraries (ROS) to develop Zutu.

The rest of the paper is as follows: in Section II, we describe Zutu's architecture and design. In section III we discuss the software stack and algorithms. In section IV we discuss the ability to scale the system. In section V we discuss the capabilities and applications of Zutu, finally, we conclude and discuss the directions for future work in section VI.

## II. PLATFORM ARCHITECTURE AND DESIGN

The design and implementation of Zutu were done using only crucial components. Hence we eliminated anything that could lead to a cost increase, keeping in mind that our platform is easily replicable, scalable, and modular. To reach the design target, we explored microcontrollers (Raspberry Pi zero, Arduino Nano, ESP8266, and ESP32), motors (DC motors with encoders, Stepper motors, and Servo motors), Wireless communication technologies (Wifi, Radio, and Bluetooth), and model designs for 3D printing. We finalised the microcontroller as ESP32 because of the quick setup times, a high number of GPIO pins, and WiFi capabilities at 2.4 GHz. We chose Wifi for wireless communication because of the minimum costs, easy accessibility, and low noise in data, lastly we decided upon using DC motors with encoders and gearboxes to provide sufficient torque. DC motors are comparatively low in cost as well as require less power compared to a Stepper motor.

### A. Hardware Design

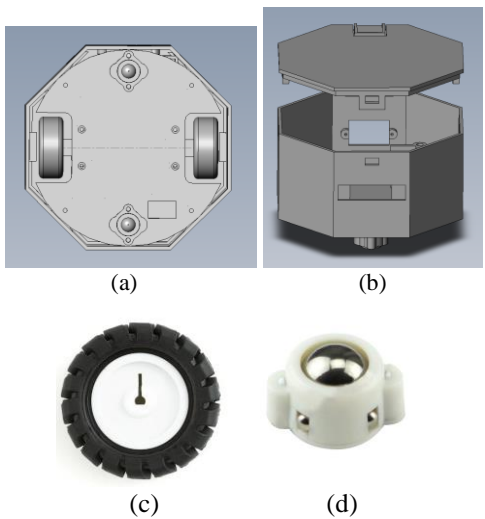

Fig. 2: Zutu 3D printed chassis and parts (a) Bottom view. (b) Front view. (c) Cylindrical wheel. (d) Castor wheel

The whole robot chassis and the top are 3D printed (Fig. 2a and 2b). Cylindrical wheels (Fig. 2c) are fitted inside to prevent anything from getting stuck into the motor shaft.

Our robots consist of 4 wheels 2 cylindrical and 2 casters (Fig. 2a) fitted precisely inside a hexagonal design which helps the robot to turn at any degree at any place in its own axis with precision. Out of four wheels, two are powered by motors and the rest two are momentum driven. DC motors are used with a gearbox which gives enough torque and speed to the wheels to make the robots move around in the environment.

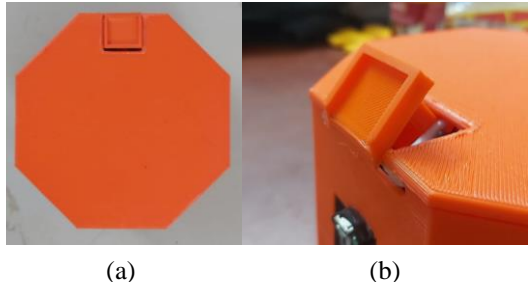

Fig. 3: Zutu prototype (a) Top view. (b) Parcel tray.

The flat top of the robot (Fig. 3a) provides sufficient space for pasting reasonably sized AR tags. In addition, there is a small parcel tray (Fig. 3b) that can contain cargo and is equipped with a precise unloading mechanism.

### B. Electronics

Motors used are 12v DC fitted with encoders which are responsible for giving the odometry data of the robot. The payload-dropping mechanism is equipped with a high-precision SG90 servo motor which precisely drops the payload at a specific position. ESP32 enables the robot to connect to LAN via Wifi, hence the robot is able to communicate with the central server (Section II(D)). Robots are fitted with a Battery level indicator which displays the status of the battery all the time and a custom LED indicator strip is also used in front of the robot which shows the current working state of the robot.

L298 is a dual motor driver, that drives the DC motors and the whole robotic system is powered by an 11.1v Li-ion battery pack of 2200 mAh which gives around 4 hours of running time to the robots.

The Hardware architecture of the individual robot is displayed (Fig. 4).

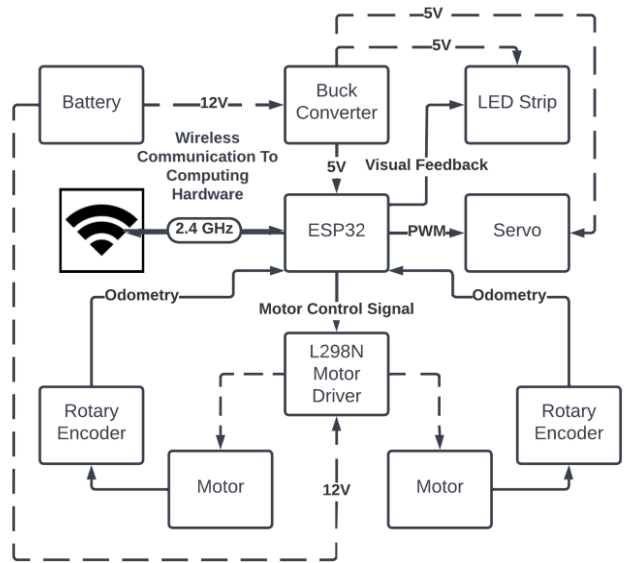

Fig. 4: Hardware Architecture of Individual robot

## C. Camera Platform for Object Tracking

The camera used is an ordinary smartphone camera (1920 x 1080 resolution) acting as an Ip camera (Fig. 5a) by transmitting the real-time feed over the network to the computation hardware. The camera is mounted over a tripod stand, viewing the working area from an angle (Fig. 5b).

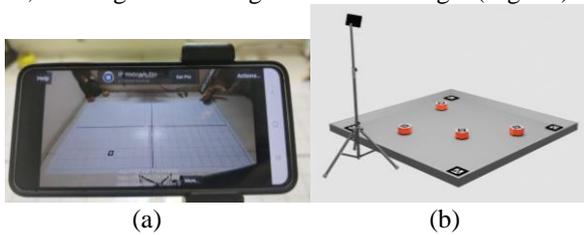

Fig. 5: (a) Camera perspective.
(b) Test setup with a camera and multiple robots

## D. Central Computing Hardware

The central server (a laptop) is the primary computation hardware running on Linux (Ubuntu 18.04), with ROS (Robot operating system) installed with the required libraries. This server is responsible for enabling communication with the robots and commanding them, getting the real-time feed from cameras and processing it to find the AR tags and extract the DoF pose of every robot. Another job of the central server is to plan the path of every single robot to avoid collisions with one another. Along with this our server also takes care of error handling like connection loss or hardware failure.

## E. Network Device

For creating a local area network home wifi router (2.4 GHz) is used with minimal configuration. Static IP addresses are assigned to each and every robot so IP addresses don't change when connections are re-established.

## III. SOFTWARE STACK AND ALGORITHMS

The software development of Zutu is done keeping in mind that the system is modular and scalable while simultaneously making the best use of the hardware components. The software stack makes use of ROS, being open source ROS has a huge database of packages that can be easily integrated with projects.

Most swarm robotics research has individual robots which are sufficiently intelligent to avoid other robots and strive to reach the objective position in the shortest amount of time. One will be able to add more robots to the system without much difficulty thanks to the simplicity of this strategy. However, due to the unpredictability of the system, we may not be able to anticipate how long it will take for the robots to reach their destination. In addition to this, there will be difficulty in calculating the system's overall performance. Similarly to our everyday world, traffic congestion is another probable system-related issue.

A much simpler option would be to dumb down the robots and implement a central traffic management system that controls the motion of each robot so as to maximize the system's overall performance.

To begin developing software for the aforementioned strategy, it is necessary to comprehend the system's fundamental requirements.

1. First, it is necessary to know where the robots are at any given moment, i.e. the position and orientation of all robots relative to a common reference.
2. Second, each robot must be governed by a software module determining its destination or movement to enable smooth motion.
3. Third is to have a control system at the robot level that accurately translates commands into the desired movements.

### A. Robot Localization: First Requirement

There are numerous methods for localising robots in the environment. One of them is using exteroceptive sensors such as a laser sensor, vision sensor, ultrasonic sensor, or lidar. Using these sensors on every robot and localising it by observing its surroundings appears doable at first glance, but it increases the cost per robot.

Another well-known way to localise is by using an IMU sensor using an accelerometer by double integration method [12]. Even though cost here is not a problem, this method comes with its own issues of the system acquiring errors over time and some type of calibration is also required.

Using computer vision, we can utilise a camera to identify robots and translate the detections into a six-degrees-of-freedom (6 DoF) Pose with position coordinates and orientation. The benefit of using this approach is that it is fairly inexpensive as no major exteroceptive sensor is needed on an individual robot.

As a continuation of the above method, it is preferable to employ a detectable tag or card to facilitate the detection of robots. Detection of augmented reality (AR) tags is computationally inexpensive and may be easily performed in real-time using computer vision. Using the camera matrix, we derive the relative posture of the AR tag with respect to the camera. If the camera's position in the surroundings is known, we can simply determine the position and orientation of the AR tag in three-dimensional space.

Multiple AR tags whose pose vectors relative to the primary tag are known can be utilised as a group to improve the overall system's accuracy. This makes the system more robust, as even if one of the AR tags is occluded, the system will continue to function with the remaining AR tags.

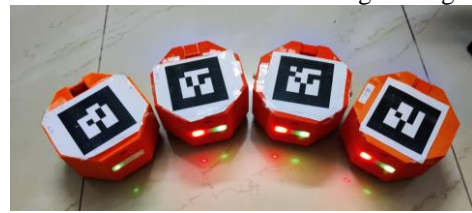

Fig. 6: Zutu robots with AR tags

A reference point is required to specify the pose (position and orientation) of an object in space. In our scenario, any fixed object may have functioned as a point of reference. One obvious reference point is the camera location.

Using AR tags on robots (Fig. 6) with a stationary central camera allows us to determine the robot's pose relative to the camera. However, this arrangement has two significant shortcomings:

1. There's an assumption that the camera will always be fixed and the whole arena is in the eye zone of the camera.

2. If the camera moves or tilts by any degree by any chance our whole system will start acting around the new errored reference point.

The solution to the above-mentioned issues is to have a relative reference point on the ground (Fig. 7). For this we fix an AR tag anywhere on the ground.

Now we have two 3D vectors first one is the one pointing to an object i.e, robot ($\overrightarrow{v_{robot}}$), and the second one is the one pointing to our reference point on the ground ($\overrightarrow{v_{ground}}$). Now using simple vector addition (1) we can easily compute the relative vector of our robot w.r.t. the ground reference point ($\overrightarrow{v_{rel}}$).

$$\overrightarrow{v_{rel}} = \overrightarrow{v_{robot}} - \overrightarrow{v_{ground}} \quad (1)$$

Now even if we move the camera in real-time making sure that both object and reference points are in the camera frame, our system will continue to work without getting affected.

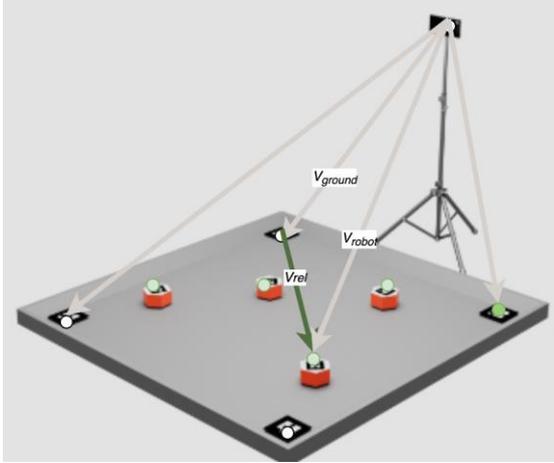

Fig. 7: Relative vector calculation

### B. Planning Module: Second Requirement

Now that we know where each robot is at any given instant, we can plan the movement of each robot using and selecting from a variety of path-planning algorithms.

To better visualise, we superimpose a virtual grid on the camera feed (Fig. 8a) to see how the system maps the plane using virtual grids. The Primary AR tag represented with a green dot (Fig. 8a) acts as the origin (0,0). We utilise the coordinates relative to the origin to determine where the robot resides within the grid. The robots are then given coordinates as waypoints to reach an end goal. This leads to great simplification in path planning.

The virtual grids are drawn over the camera feed by taking reference from the AR tag fitted on the ground and then calculating the coordinates of the grid to cover the workspace. These physical parameters such as the size of the grid and the distance between two coordinates are then operated with the reverse camera matrix to translate these coordinates into the camera frame and then drawn back on the live feed with the OpenCV library [13] functions.

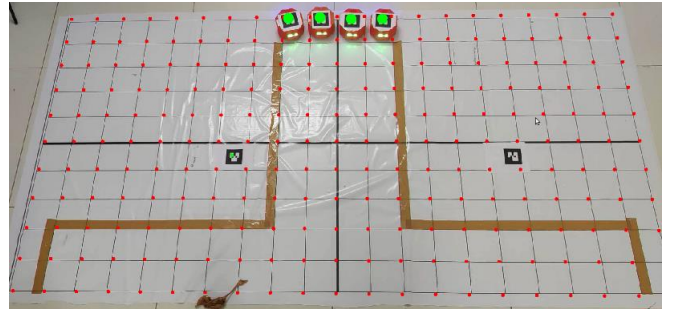

(a)

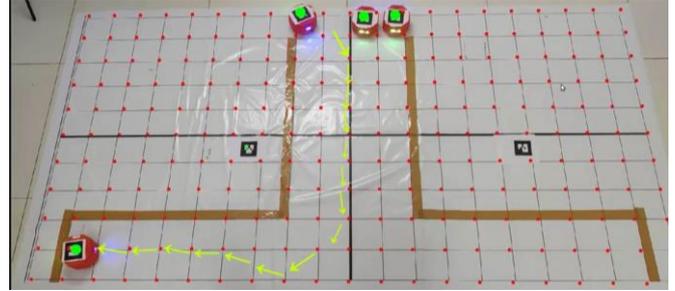

(b)

Fig. 8: (a) Virtual grid overlaid on the physical grid
(b) Zutu robots being able to reach the final goal by following the plan in the grid

By setting basic criteria in the planning algorithm, such as no two robots being in the same grid at the same time, collisions with other robots are prevented.

The output of the planning algorithm (Fig. 8b) is a list of waypoints for each robot indicating where it should be at the $n^{th}$ instant of time. Once the list of waypoints is complete, all that remains is to command the robots to execute their separate plans and move. A video accompanying this scenario is available at https://youtu.be/ESO9nx7IlDA.

As there will be instances in which the final destination will be far from the starting point and the system may experience slippage, we chose to employ the waypoint strategy to reduce cumulative error.

In Fig. 9, P1 was the initial waypoint, but owing to some error (due to slipping), it reached R1. P2 was the next waypoint, but due to an error, it reached R2. And then it advances towards the final point and eventually reaches there. Adding waypoints helps the robot acquire fewer errors in position and direction. The frequency of the waypoints can be a function of the distance between the start and endpoints.

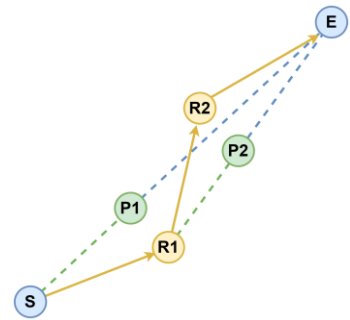

Fig. 9: Waypoint navigation

To better estimate the performance of the planning algorithm and analyse the motion during debugging and answering questions such as where the robot is and how accurate its motion was without measuring with physical metrics, it is very useful to print the virtual grid on the live camera feed as it is not possible to physically place these dots or grids on the ground solely for the purpose of analysis.

Virtual grids are also useful in the early setup of the project, as they aid in locating the appropriate placements for reference augmented reality (AR) tags on the ground in order to best cover the workspace and determine the optimal camera angle.

*C. Controlling The Robot: Third Requirement*

Input for the robot is a position where it should be present at the next instant of time (Fig. 10). Considering point A to be the present position and B to be the goal position. The movement from point A to point B can be described as two simple steps.

1. Align in the direction of goal
2. Move towards the goal

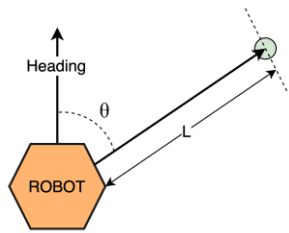

Fig. 10: Robot waypoint navigation

The last step is to translate the motion into the number of turns of the wheels, followed by the number of steps the encoder must observe to obtain the desired final state, which results in the required motion.

Each robot is equipped with a programmed ESP32 module that houses the entire lower-level PID control system; encoders provide step-based feedback. PID control loop (Fig.11) is accountable for receiving the objective as the number of steps for each motor and getting the desired step configuration in the shortest time conceivable. Since all robots are physically identical, PID constants are manually tweaked using the graph approach and reproduced to additional robots with minor modifications.

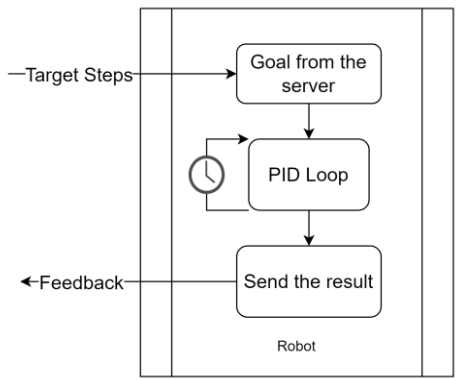

Fig. 11: PID loop in the individual robots

## IV. SCALING THE SYSTEM

The software architecture (Fig. 12) allows the easy addition of robots without many changes in the core system. To add a robot one only needs to assign a static IP address to the robot and tell the central server of an additional robot in the network.

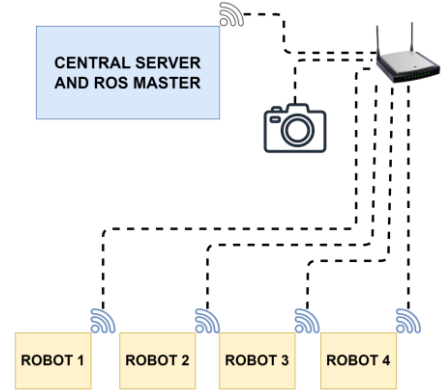

Fig. 12: Software Architecture

Our experiments involved four robots with the configurations listed in Table I.

TABLE I      ROUTER CONFIGURATIONS

| Router Configuration | | |
|---|---|---|
| Sr. No. | Node Name | IP Address |
| 1. | Router | 192.168.1.1 |
| 2. | Central Server (Laptop) | 192.168.1.47 |
| 3. | IP Camera 1 | 192.168.1.25 |
| 4. | Robot 1 | 192.168.1.4 |
| 5. | Robot 2 | 192.168.1.5 |
| 6. | Robot 3 | 192.168.1.6 |
| 7. | Robot 4 | 192.168.1.7 |

## V. CAPABILITIES AND APPLICATIONS

After describing the software and hardware stack of our platform, we did some experimentation to describe the capabilities of our solution.

*A. Navigation*

In Zutu, a goal coordinate can be specified in cartesian coordinates, and the robot can reach the specified objective by computing the waypoints along the journey. The distances between waypoints can be modified to improve the fluidity of the robots' movement. This can be utilised to drop packages placed atop robots at a specified coordinate. In a warehouse, the platform can be used to sort packages.

*B. Resilience*

The platform is resistant to environmental changes and does not require regular sensor calibrations. Based on the stationary position of the AR tag on the ground, the camera identifies the robots' relative positions. Therefore, as long as the ground AR tags remain fixed, the generated virtual grid will not move, and the robots can operate as expected in the physical workspace.

*C. Multi-robot path planning*

The principal logic of the platform lies in the central server (computational hardware). Hence path planning algorithms can be integrated that work with cartesian coordinates to navigate the robots dynamically to reach the goal while avoiding other robots' planned paths. Such testing feasibility can be used for research and educational purposes to facilitate the quick deployment of path-planning algorithms. For example, Multirobot path planning using Graphs [14].

*D. Scaling to increase workspace area*

To scale the system, just add more cameras and secondary reference points (Fig 12a), ensuring that at least one reference point is inside the overlapping field of view of the cameras. The AR tag fitted on the ground will serve as the secondary reference point and help the two systems link to the primary reference point. Then, using simple vector addition (Fig 12b), we obtain the robot's resultant vector.

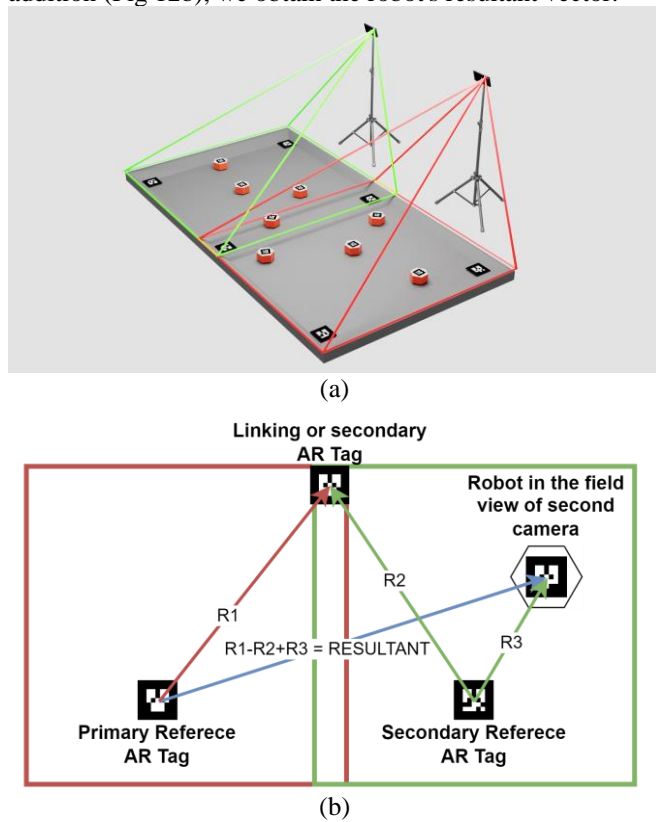

Fig. 12: (a) Multiple camera setup to scale Zutu
(b) Vector representation using multiple camera setup

## VI. CONCLUSION AND FUTURE WORK

In this paper we presented Zutu, a low-cost, modular, and easily scalable platform for research and educational purposes to run swarm robots using a monocular camera and no onboard sensors for research and educational purposes.

Our work also examined the resistance of our localization technique to obstructions and camera disturbance. Using our system, the camera can be moved at any time, and our system will adapt to the new position. Adding supplementary reference points to the ground (AR tags) and combining numerous cameras can also extend the area covered.

In our future work, we intend to expand Zutu by incorporating appropriate multi-robot path planning algorithms in a cartesian plane to reach the end point from the start coordinate while avoiding other robots. We also intend to improve the fluidity between waypoints.